%% file: main.tex
\documentclass[11pt]{article}

\PassOptionsToPackage{table,dvipsnames}{xcolor}
\usepackage[preprint]{acl}

\usepackage{times}
\usepackage{latexsym}

\usepackage[T1]{fontenc}
\usepackage[utf8]{inputenc}

\usepackage{microtype}

\usepackage{inconsolata}

\usepackage{graphicx}

\usepackage{amsmath}
\usepackage{amssymb}
\usepackage{mathtools}
\usepackage{amsthm}

\usepackage[capitalise,noabbrev]{cleveref}

\usepackage{subcaption}
\usepackage{booktabs} % for professional tables
\usepackage{xcolor}
\usepackage{bm}
\usepackage{multirow}
\usepackage{array}
\usepackage{pifont}
\usepackage{enumitem}
\usepackage{listings}

\usepackage{algorithm}
\usepackage{algpseudocode}

\let\INPUT\Require
\let\OUTPUT\Ensure
\let\STATE\State
\let\FOR\For
\let\ENDFOR\EndFor
\let\WHILE\While
\let\ENDWHILE\EndWhile

\newcommand{\ourmethod}{{AWARe}}

\newcommand{\zeroshot}{{Zero-shot}}
\newcommand{\fullft}{{Full-FT}}
\newcommand{\lora}{{LoRA}}
\newcommand{\dora}{{DoRA}}
\newcommand{\orthreg}{{Orth-Reg}}
\newcommand{\dare}{{DARE}}
\newcommand{\tailor}{{Model Tailor}}
\newcommand{\lorasculpt}{{LoRASculpt}}
\newcommand{\spider}{{SPIDER}}

\newcommand{\iconqa}{{IconQA}}
\newcommand{\coco}{{COCO-Caption}}
\newcommand{\okvqa}{{OKVQA}}
\newcommand{\ocrvqa}{{OCRVQA}}
\newcommand{\gqa}{{GQA}}
\newcommand{\textvqa}{{TextVQA}}

\definecolor{DarkBlue}{RGB}{64,101,149}
\definecolor{azure}{rgb}{0.0, 0.5, 1.0}
\definecolor{gray}{rgb}{0.3, 0.3, 0.3}
\definecolor{DarkGreen}{RGB}{42,110,63}
\definecolor{DarkYellow}{RGB}{191,144,0}
\definecolor{DarkRed}{rgb}{0.6, 0, 0}
\definecolor{LightOrange}{RGB}{246, 198, 172}
\definecolor{OceanBlue}{RGB}{171, 224, 240}

\newcommand{\greenup}[1]{$_{\color{green}\uparrow #1}$}

\newcommand{\thickhline}{\noalign{\hrule height 1.2pt}}
\newcolumntype{x}[1]{>{\centering\arraybackslash}p{#1pt}}
\newcolumntype{I}{!{\vrule width 1pt}}

\newcommand{\pub}[1]{{\color{gray}{\footnotesize{[{#1}]}}}}

\newenvironment{fullitemize}
{
\vspace{-1pt}
\begin{itemize}[leftmargin=*]
\setlength{\itemsep}{5pt}
\setlength{\parsep}{-5pt}
\setlength{\parskip}{-3pt}
\setlength{\leftmargin}{-10pt}
}
{
\end{itemize}
\vspace{-1pt}
}

\graphicspath{ {./figures/} }

\theoremstyle{plain}

\theoremstyle{definition}

\theoremstyle{remark}

\title{\raisebox{-.15\height}{\includegraphics[height=1em]{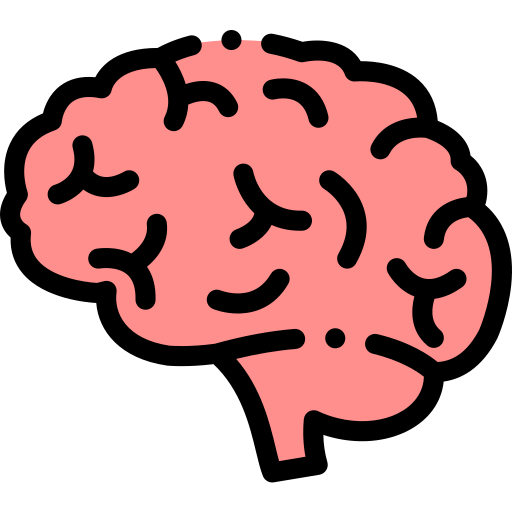}} AWARe: Mitigating Catastrophic Forgetting via Activation-Weighted Adaptive REtention}

\author{
  \textbf{Juncheng Liao\textsuperscript{1,*}},
  \textbf{Jinfan Lv\textsuperscript{2,*}},
  \textbf{Guoming Wang\textsuperscript{1,$\dagger$}},
  \\
  \textbf{Jupeng Zheng\textsuperscript{3}},
  \textbf{Ling Xiao\textsuperscript{4}},
  \textbf{Siliang Tang\textsuperscript{1}}
  \\
  \\
  \textsuperscript{1}School of Software Technology, Zhejiang University, Hangzhou, China \\
  \textsuperscript{2}College of Intelligent Robotics and Advanced Manufacturing, Fudan University, Shanghai, China \\
  \textsuperscript{3}School of Artificial Intelligence, Sun Yat-Sen University, Guangzhou, China \\
  \textsuperscript{4}Graduate School of Information Science, Hokkaido University, Sapporo, Japan \\
  \small{
    \textsuperscript{*}\textbf{Equal contribution.}\quad
    \textsuperscript{$\dagger$}\textbf{Correspondence:}
    \href{mailto:NB21013@zju.edu.cn}{NB21013@zju.edu.cn}
  }
}

\begin{document}
\maketitle
\begin{abstract}
\input{sections/0_Abstract}
\end{abstract}

\section{Introduction}
\input{sections/1_Introduction}

\section{Related Work}
\input{sections/2_Related_Work}

\section{Methodology}
\input{sections/3_Methodology}

\section{Experiments}
\input{sections/4_Experiments}

\section{Conclusion}
\input{sections/5_Conclusion}

\section{Limitations}

AWARe relies on a calibration set to estimate activation-based neuron importance. Although our experiments show that a general-purpose dataset such as MMMU can provide a practical substitute when upstream task data is unavailable, the quality and coverage of the calibration data may still affect the selected frozen regions. In addition, our evaluation focuses on representative MLLM backbones and multimodal tasks; broader validation on larger models, more diverse domains, and longer continual learning sequences would further clarify the generality of the approach. Finally, because AWARe explicitly freezes salient parameters to preserve prior capabilities, extremely large distribution shifts may still require tuning the retention ratio to balance stability and plasticity.

\nocite{*}

\bibliography{references}

\appendix

\include{sections/Appendix}

\end{document}

%% file: sections/0_Abstract.tex
Multimodal Large Language Models (MLLMs) exhibit strong generalization and reasoning abilities due to large-scale multimodal pre-training. However, fine-tuning these models on downstream tasks often leads to catastrophic forgetting, where newly learned task-specific knowledge degrades previously acquired capabilities. This issue arises because gradient updates for new tasks overwrite parameters critical to prior knowledge, limiting the practical deployment of MLLMs.
To address this challenge, we propose \textbf{Activation-Weighted Adaptive REtention (AWARe)}, a fine-tuning method that mitigates catastrophic forgetting by dynamically controlling parameter updates based on activation patterns. AWARe assigns activation-based importance scores to parameters, selectively freezing those essential for preserving prior capabilities while allowing less important parameters to adapt to new tasks. Importantly, AWARe operates \textit{without modifying model architectures}, ensuring compatibility with existing inference engines. Extensive experiments demonstrate that AWARe effectively preserves upstream capabilities while achieving superior downstream performance compared to existing methods. Code is available at \href{https://github.com/kaln27/AWARe}{https://github.com/kaln27/AWARe}.

%% file: sections/1_Introduction.tex
\begin{figure}[t]
    \centering
    \includegraphics[width=\columnwidth]{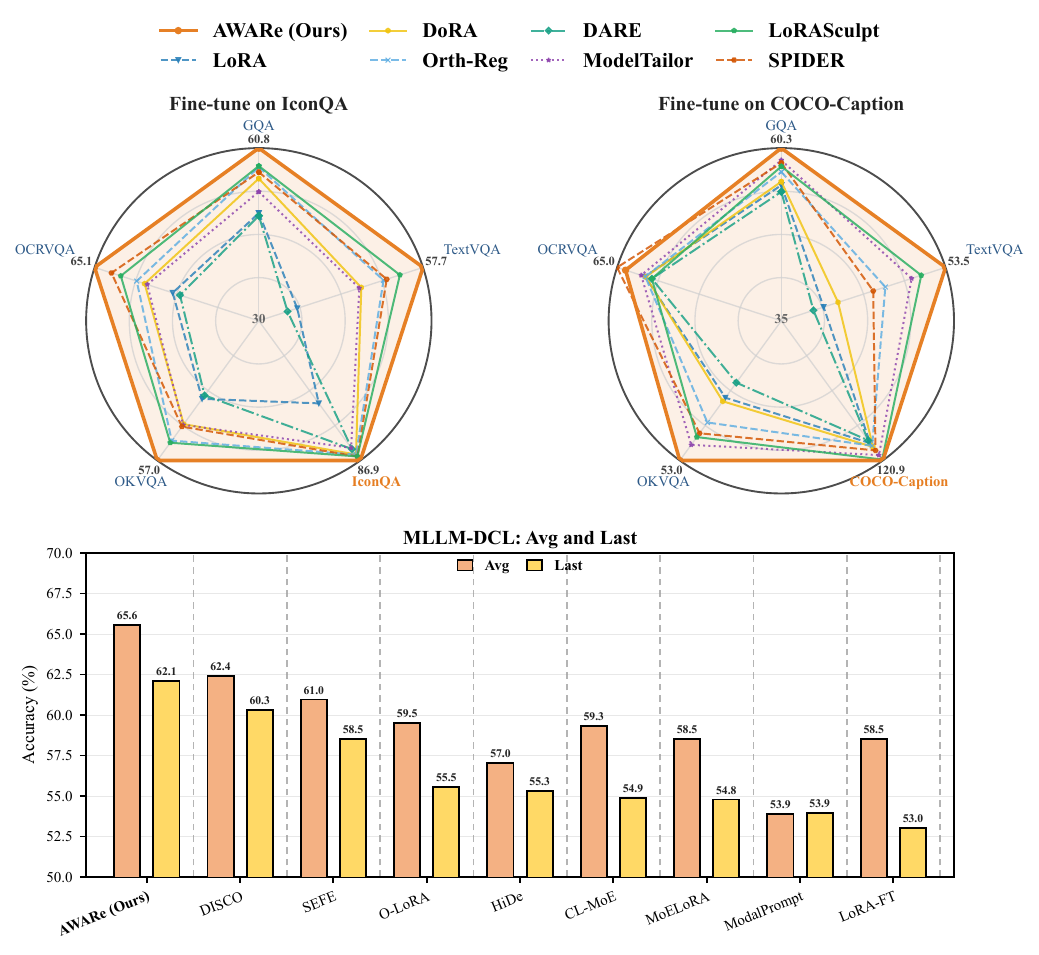}
    \caption{\textbf{Overall Performance of AWARe.} Top: radar plots showing comparison between our approach and various baselines under Single-Task Downstream Adaptation. Bottom: results on the MLLM-DCL continuous learning benchmark.}
    \label{fig:overall_perf}
    \vspace{-10pt}
\end{figure}

\begin{figure*}[t]
    \centering
    \includegraphics[width=0.9\textwidth]{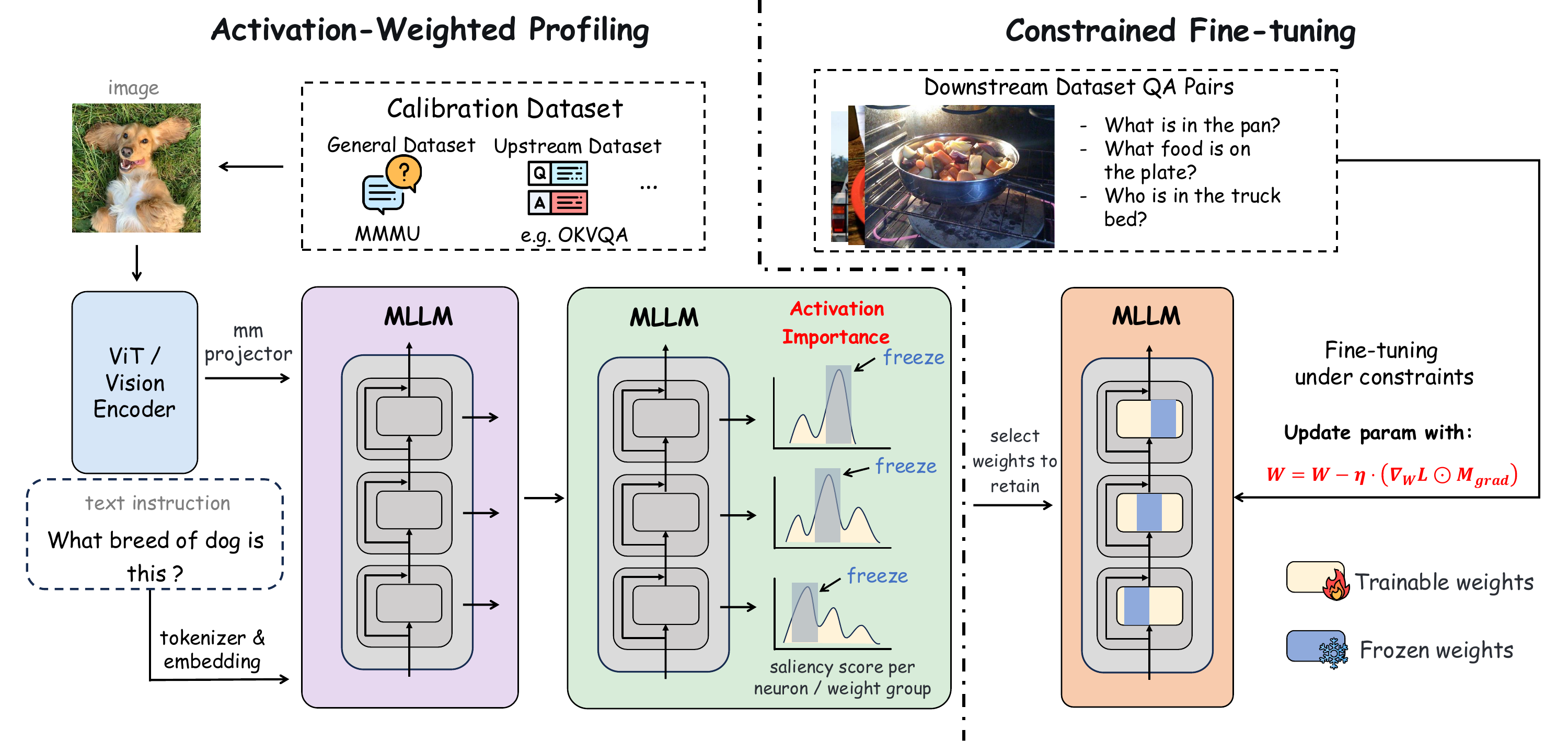}
    \caption{\textbf{Overview of AWARe.} Given a multimodal base model, AWARe identifies important neurons or weight groups using activation-based saliency on calibration samples, then freezes high-saliency weights and updates only low-saliency weights during fine-tuning.}
    \label{fig:method_overview}
    \vspace{-10pt}
\end{figure*}

Multimodal Large Language Models (MLLMs) have emerged as powerful versatile agents, capable of solving complex tasks that require understanding and reasoning across both inputs of vision and language \cite{zhu2023minigpt,InstructBLIP_NeurIPS23,QwenVL_arXiv23,LLaVA15_CVPR24,VILA_CVPR24,chen2024internvl,Qwen2VL_arXiv24,zhu2025internvl3,wang2025internvl3_5,Qwen2_5VL_arXiv25,Qwen3VL_arXiv25}. 
Typical MLLMs are composed of a pre-trained visual encoder, a large language model (LLM), and a connector module \cite{BLIP_ICML22,InstructBLIP_NeurIPS23,LLaVA15_CVPR24}. These models facilitate strong generalization abilities acquired from large-scale pre-training. 
Despite their impressive zero-shot capabilities, fine-tuning MLLMs on downstream tasks remains a standard practice to tailor models for specific domains or improve instruction-following performance \cite{InstructBLIP_NeurIPS23,QwenVL_arXiv23,LLaVA15_CVPR24}. 
However, this adaptation process often comes at a cost: \textit{catastrophic forgetting} (CF), where the model's proficiency in previously learned upstream tasks degrades significantly as it optimizes for new objectives \cite{mccloskey1989catastrophic,mcclelland1995there,kirkpatrick2017overcoming,li2024revisiting}. 
Nonetheless, this form of forgetting remains particularly severe in MLLMs due to the complex interplay between modalities and the high dimensionality of the parameter space \cite{sung2023ecoflap,SpeandGenonCF_arXiv23,CFforLLMinCFT_arXiv23,zhai2024investigating,shen2024multimodal,jiang2024effectiveness,CFinMLLMFT_CPAL24}.

Existing approaches to mitigate catastrophic forgetting, such as experience replay \cite{riemer2018learning,chaudhry2019tiny} or regularization-based methods \cite{kirkpatrick2017overcoming}, often incur high computational overhead or struggle to scale to the billions of parameters in MLLMs. 
Parameter-Efficient Fine-Tuning (PEFT) methods like LoRA \cite{LoRA_ICLR22} reduce the trainable parameter count but do not inherently prevent the erosion of pre-trained knowledge, as recent studies show they can still disrupt reliable upstream features \cite{biderman2024lora,ModelTailor_ICML24}. 
Consequently, a critical challenge arises: \textit{How can we enable MLLMs to adapt effectively to downstream tasks while robustly preserving their pre-trained capabilities without prohibitive costs?}

Our work is primarily inspired by the principle of activation-aware parameter saliency, most notably demonstrated by Activation-aware Weight Quantization (AWQ) \cite{AWQ_MLSYS24}. 
AWQ shows that a small fraction of weights associated with large activation magnitudes are disproportionately important for preserving model behavior under quantization. 
Wanda reaches a complementary conclusion in pruning: static weight magnitude alone is insufficient for LLMs, while combining weights with input activation statistics gives a simple and effective estimate of functional importance \cite{WANDA_ICLR24}. 
Recent model merging studies further suggest that activation-guided consensus can reduce parameter interference when combining specialized models \cite{AIM_NeurIPS25,ACM_NeurIPS25}. 
Together, these findings indicate that activations provide a dynamic, data-dependent view of which internal pathways support stable behavior, beyond what weight-centric criteria such as magnitude pruning can capture \cite{han2015deep,LTH_ICLR19}. 
We therefore repurpose activation saliency from compression and merging to forgetting mitigation.

Building on this motivation, we introduce \textbf{AWARe}, short for \textbf{A}ctivation-\textbf{W}eighted \textbf{A}daptive \textbf{RE}tention, a novel framework designed to surgically mitigate catastrophic forgetting. 
AWARe has two phases: \textit{Knowledge Profiling}, which estimates neuron saliency from a small calibration set, and \textit{Constrained Fine-tuning}, which freezes the most salient parameters during downstream training. 
When upstream data are unavailable, we can use a general-purpose calibration set such as MMMU \cite{yue2023mmmu}; as shown by the \ourmethod{} (MMMU) results in \cref{tab:main_results}, this remains highly effective while adding only a negligible profiling cost.

We validate our approach through comprehensive experiments on IconQA \cite{IconQA_NeurIPS21}, COCO-Caption \cite{COCO_ECCV14}, and MLLM-DCL \cite{MLLM_CL_arXiv25} (Continuous learning) benchmarks.
AWARe achieves strong stability-plasticity trade-offs, and freezing the top 30\% of self-attention parameters is often sufficient to preserve upstream knowledge while keeping downstream performance competitive.

Our main contributions are summarized as follows:
\begin{fullitemize}
    \item[\ding{182}] \textbf{AWARe Framework.} We introduce Activation-Weighted Adaptive REtention (AWARe), a simple activation-based method for preserving critical upstream knowledge during downstream adaptation in MLLMs.
    
    \item[\ding{183}] \textbf{Efficiency and Simplicity.} Our approach requires no architectural changes and updates only a small subset of parameters (e.g., $\sim17.5\%$; see \cref{sec:param_analysis}), avoiding replay buffers and extra modules.
    
    \item[\ding{184}] \textbf{Comprehensive Validation.} We show through ablations and benchmarks that activation-based saliency is effective and that AWARe preserves generalization across diverse settings.
\end{fullitemize}

%% file: sections/2_Related_Work.tex
% \subsection{Multimodal Large Language Models}
% Building upon the formidable reasoning capabilities of Large Language Models (LLMs) such as LLaMA \cite{LLaMA_arXiv23, touvron2023llama} and Vicuna \cite{Vicuna_arXiv23}, the research community has witnessed a paradigm shift towards Multimodal Large Language Models (MLLMs). These systems fundamentally aim to perceive and reason about visual signals as effectively as textual data. Pioneering frameworks like LLaVA \cite{LLaVA15_CVPR24} and MiniGPT-4 \cite{zhu2023minigpt} treat visual inputs as pseudo-tokens, projecting features extracted by strong visual encoders like CLIP \cite{CLIP_ICML21} into the LLM's native embedding space. Diversity exists in the alignment mechanisms: while methodology like LLaVA adopts a concise Multi-Layer Perceptron (MLP) for projection, architectures such as BLIP-2 \cite{BLIPv2_ICML23} and InstructBLIP \cite{InstructBLIP_NeurIPS23} introduce the Query Transformer (Q-Former) to compress visual representations. Further advancements have led to models with enhanced resolution and capability, including Qwen-VL series \cite{QwenVL_arXiv23, Qwen2VL_arXiv24, Qwen2_5VL_arXiv25, Qwen3VL_arXiv25}, VILA \cite{VILA_CVPR24}, and InternVL \cite{chen2024internvl}. Crucially, the paradigm of visual instruction tuning \cite{LLaVA15_CVPR24} empowers these models to generalize across unseen tasks by training on broad instruction-following data, although adapting them to specific vertical domains often necessitates further fine-tuning.

\subsection{Catastrophic Forgetting and Continual Learning}
Fine-tuning serves as a pivotal mechanism for adapting Multimodal Large Language Models (MLLMs) to downstream tasks. However, deep learning models often suffer from catastrophic forgetting \cite{mccloskey1989catastrophic, mcclelland1995there}, where previously learned knowledge is lost when acquiring new skills. Various continual learning algorithms have been proposed to address this issue, generally encompassing rehearsal-based, regularization-based \cite{kirkpatrick2017overcoming, lopez2017gradient}, and architecture-based approaches \cite{houlsby2019parameter, lester2021power}. While established, many traditional methods rely on full-model fine-tuning or second-order statistics, making them computationally prohibitive for large-scale models.

\subsection{Parameter-Efficient Fine-Tuning for Mitigation}
In the era of large foundation models, the challenge shifts towards maintaining the model's strong generalization and "Open-World Stabilization" after downstream adaptation \cite{ni2023forgetting, zhai2024investigating}. Recent efforts have explored parameter-efficient solutions to mitigate forgetting. Reparameterization methods such as DoRA \cite{DoRA_ICML24} and LoRA-based extensions like LoRAMoE \cite{dou2024loramoe} aim to enhance learning capacity while addressing generalization loss, whereas LoRASculpt \cite{LORASCULPT_CVPR25} prunes redundant parameters via sparse updates. Regularization techniques have also been adapted; for instance, Orth-Reg \cite{Orth-Reg_ECCV24} encourages fine-tuned features to remain orthogonal to pre-trained representations. CorDA \cite{yang2024corda} builds task-aware adapters through context-guided weight decomposition to preserve world knowledge. Selective tuning approaches, such as SPIDER \cite{SPIDER_ICML25} and Model Tailor \cite{ModelTailor_ICML24}, leverage gradient or saliency analysis to construct sparse updates, aiming to protect critical pre-trained features. However, these criteria often rely on static information or gradient magnitudes, overlooking the dynamic nature of neuronal activations during forward propagation.

Recognizing these limitations, we propose AWARe. Unlike additive methods that introduce extra parameters or selective tuning that relies on static priors, AWARe utilizes task-induced activation distributions to measure parameter importance. By adaptively constraining high-activation neurons based on upstream statistics, AWARe effectively balances the preservation of core capabilities with the flexibility needed for downstream task adaptation.

%% file: sections/3_Methodology.tex
\subsection{Problem Formulation}
A Multimodal Large Language Model (MLLM) $\mathcal{M}_\theta$ generally comprises three core components: a vision encoder $\mathcal{V}$ (e.g., CLIP \cite{CLIP_ICML21} or SigLIP \cite{zhai2023sigmoid}), a large language model $\mathcal{L}$ (e.g., LLaMA \cite{touvron2023llama,LLaMA_arXiv23} or Vicuna \cite{Vicuna_arXiv23}), and a connector module $\rho$ that bridges the visual and textual modalities. The standard paradigm involves aligning the pre-trained vision encoder representations with the LLM's embedding space via the connector, enabling the model to process multimodal inputs effectively.

\begin{figure}[h]
    \centering
    \includegraphics[width=\columnwidth]{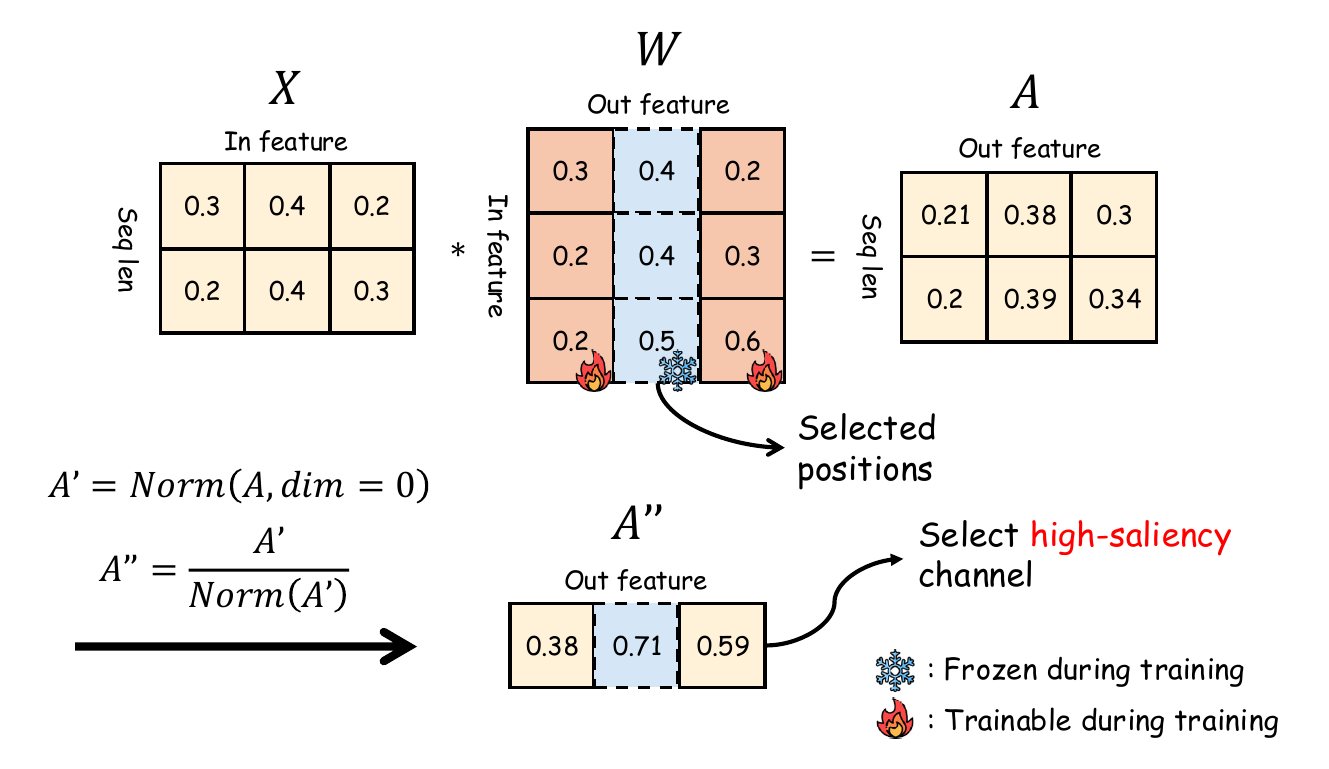}
    \caption{Overview of AWARe. For each target linear layer, we profile the activation matrix produced by calibration samples, normalize the activation statistics across sequence and feature dimensions, and obtain a saliency score for each output channel. High-saliency channels are selected and frozen to preserve pretrained knowledge, while the remaining channels stay trainable for downstream adaptation.}
    \label{fig:method}
    \vspace{-10pt}
\end{figure}

Consider a Multimodal Large Language Model (MLLM) $\mathcal{M}_\theta$ parameterized by $\theta$. We assume the model has been initially trained on a foundational set of upstream tasks $\mathcal{T}_{base}$. In a continual instruction tuning setting, the model is sequentially exposed to a stream of downstream tasks, denoted as $\mathcal{T}_{seq} = \{\mathcal{T}_1, \mathcal{T}_2, \dots, \mathcal{T}_N\}$. Our objective is to incrementally learn each new task $\mathcal{T}_t$ at step $t$ by updating $\theta$, while mitigating the \textit{catastrophic forgetting} \cite{mccloskey1989catastrophic,mcclelland1995there} of both the foundational capabilities from $\mathcal{T}_{base}$ and the previously acquired knowledge from tasks $\{\mathcal{T}_1, \dots, \mathcal{T}_{t-1}\}$. As illustrated in \cref{fig:method}, we achieve this by partitioning the parameters into a fixed set $\theta_{fixed}$, which captures and preserves core established features, and a plastic set $\theta_{active}$ used for acquiring new task-specific knowledge.

\subsection{Activation-Based Saliency Estimation}
To identify important parameters for retention, we utilize a calibration set $\mathcal{S}_{cal}$ composed of representative samples from each upstream task's train datasets ($\mathcal{T}_{up}$). Specifically, we randomly sample several instances from each upstream task. In scenarios where access to upstream data is restricted, we demonstrate that utilizing a comprehensive general benchmark, such as MMMU \cite{yue2023mmmu}, as a proxy for $\mathcal{S}_{cal}$ is equally effective. We empirically demonstrate the robustness of model performance to the calibration set size in our ablation studies at \cref{sec:calibration_dataset_sensitivity}. Crucially, the profiling incurs negligible overhead, requiring only a single forward pass on a small sample set. For a target linear layer with weight matrix $\mathbf{W} \in \mathbb{R}^{d_{out} \times d_{in}}$, we analyze its activation tensor $\mathbf{A} \in \mathbb{R}^{B \times L \times d_{out}}$, where $i \in \{1, \dots, B\}$, $j \in \{1, \dots, L\}$, and $k \in \{1, \dots, d_{out}\}$ index the batch size, context length and hidden dimension, respectively. 

The saliency estimation follows a three-step aggregation and normalization process. First, we compute the $L_2$-norm of activations along the sequence length dimension for each sample $j$ and neuron $k$:
\begin{equation}
a'_{i,k} = \sqrt{\textstyle\sum_{j=1}^{L} a_{i,j,k}^2}
\label{eq:norm}
\end{equation}
Second, to ensure comparability across samples and prevent those with outlier-scale activations from biasing the estimations, we apply per-sample $L_2$-normalization across the hidden dimension:
\begin{equation}
a''_{i,k} = \frac{a'_{i,k}}{\|\mathbf{a}'_i\|_2} = \frac{a'_{i,k}}{\sqrt{\textstyle\sum_{k'=1}^{d_{out}} {a'_{i,k'}}^2}}
\label{eq:normalize}
\end{equation}
where $\mathbf{a}'_i = [a'_{i,1}, \dots, a'_{i,d_{out}}]^\top$ is the vector of sequence-aggregated activations for sample $i$. This normalization ensures that the saliency is determined by the \textit{relative} importance of neurons within each context rather than absolute activation scale, which is crucial for identifying neurons that consistently capture structural features across diverse upstream tasks. This refinement distinguishes our approach from simple magnitude-based methods and explains the superior performance observed in \cref{tab:ablation_selection}.
Finally, the saliency score for each neuron $k$ is obtained by averaging across the batch:
\begin{equation}
s_k = \frac{1}{B} \textstyle\sum_{i=1}^{B} a''_{i,k}
\label{eq:saliency}
\end{equation}
The resulting saliency vector $\mathbf{s} \in \mathbb{R}^{d_{out}}$ reflects the relative importance of each output neuron in preserving upstream task capabilities.

\subsection{Adaptive Parameter Retention}
Guided by the saliency scores, we introduce a retention ratio $\rho \in (0, 1)$ as a hyperparameter to control the fraction of parameters to be frozen. We rank output neurons across all target linear layers by their saliency scores and identify the global index set $\mathcal{I}$ containing the top-$\rho$ fraction of neurons (e.g., top $30\%$ or $10\%$).

\begin{figure*}[t]
    \centering
    \begin{subfigure}[b]{0.32\textwidth}
        \centering
        \includegraphics[width=\textwidth]{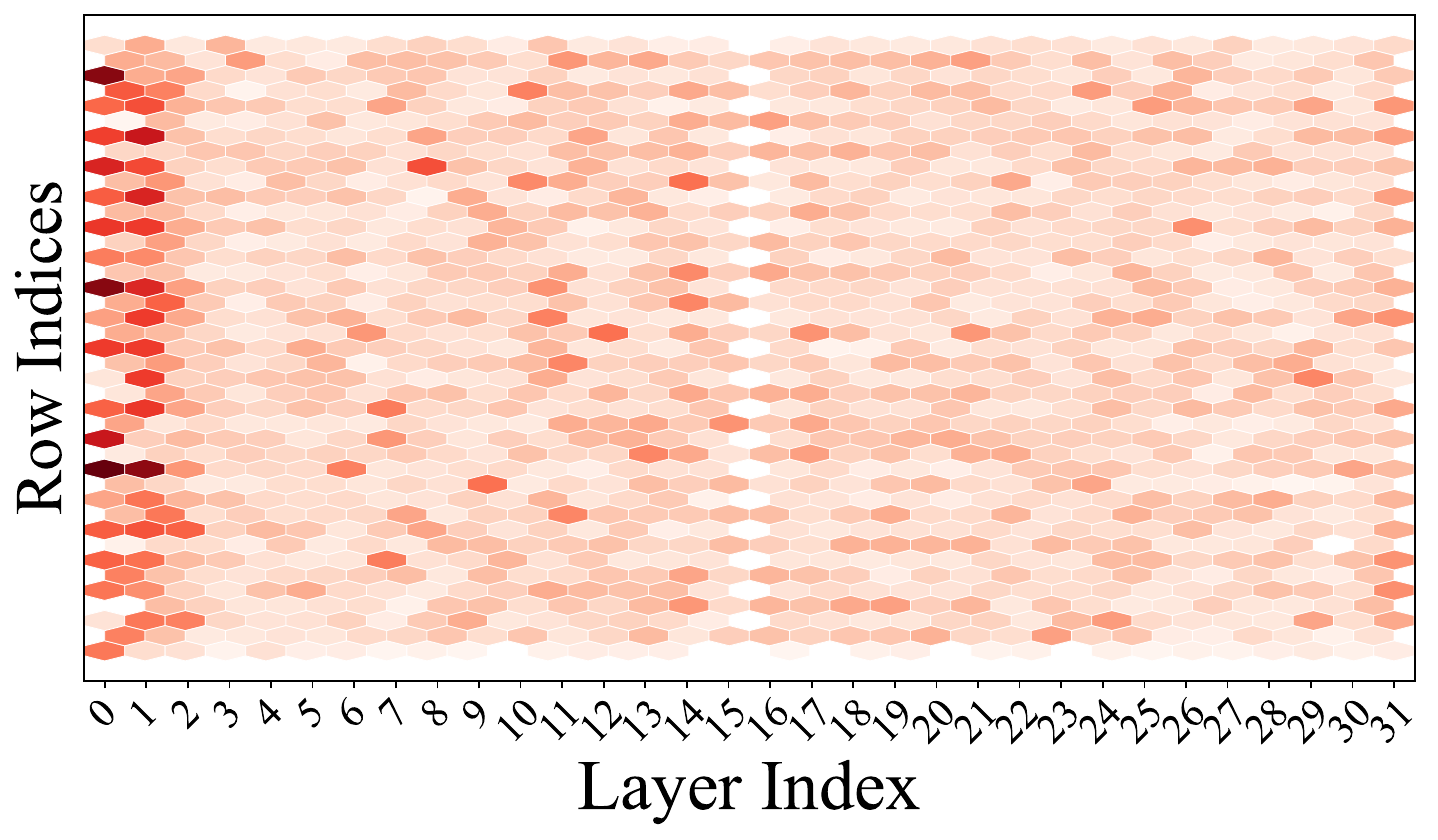}
        \caption{\texttt{q\_proj}}
        \label{fig:q_proj}
    \end{subfigure}
    \hfill
    \begin{subfigure}[b]{0.32\textwidth}
        \centering
        \includegraphics[width=\textwidth]{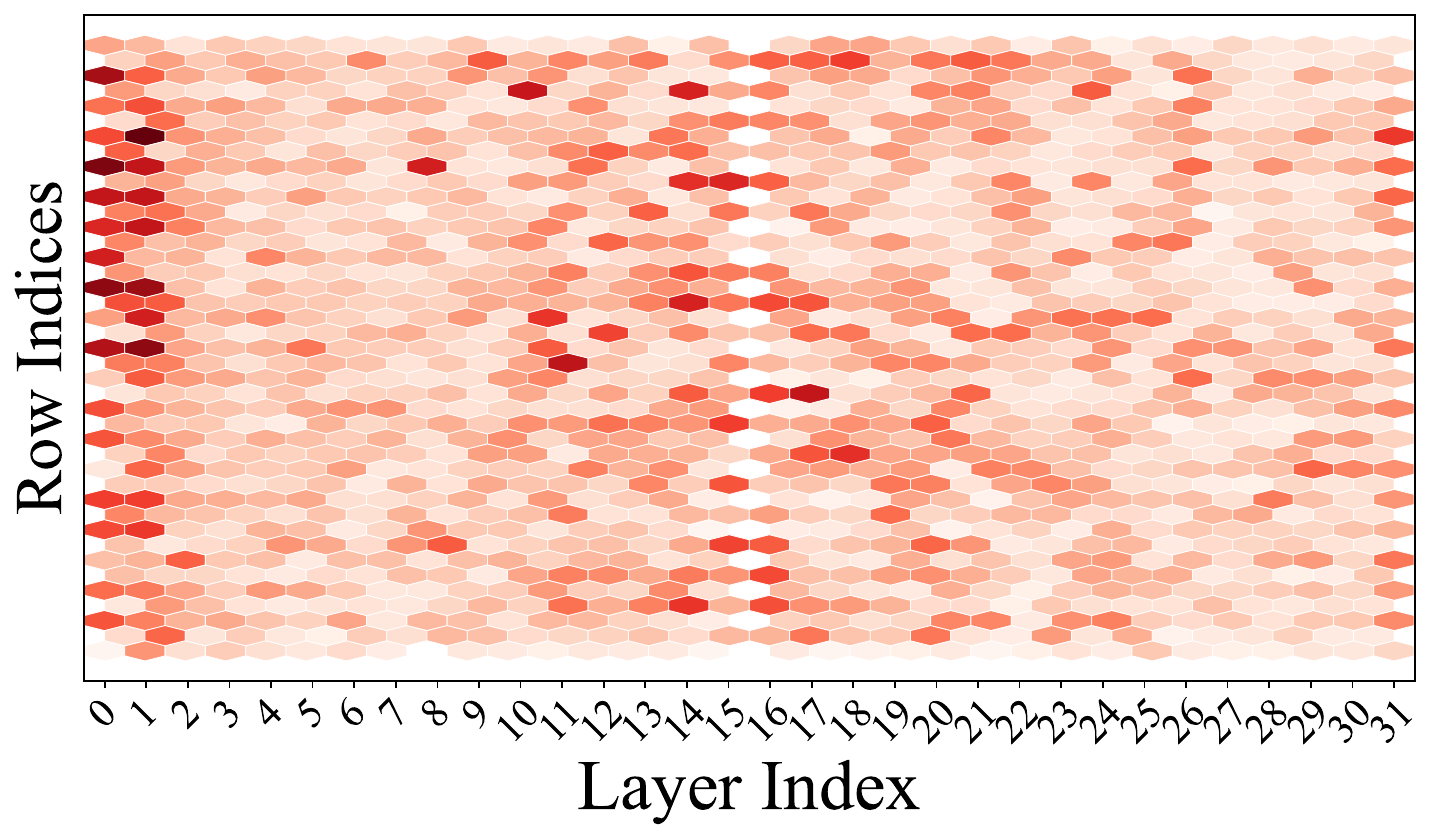}
        \caption{\texttt{k\_proj}}
        \label{fig:k_proj}
    \end{subfigure}
    \hfill
    \begin{subfigure}[b]{0.32\textwidth}
        \centering
        \includegraphics[width=\textwidth]{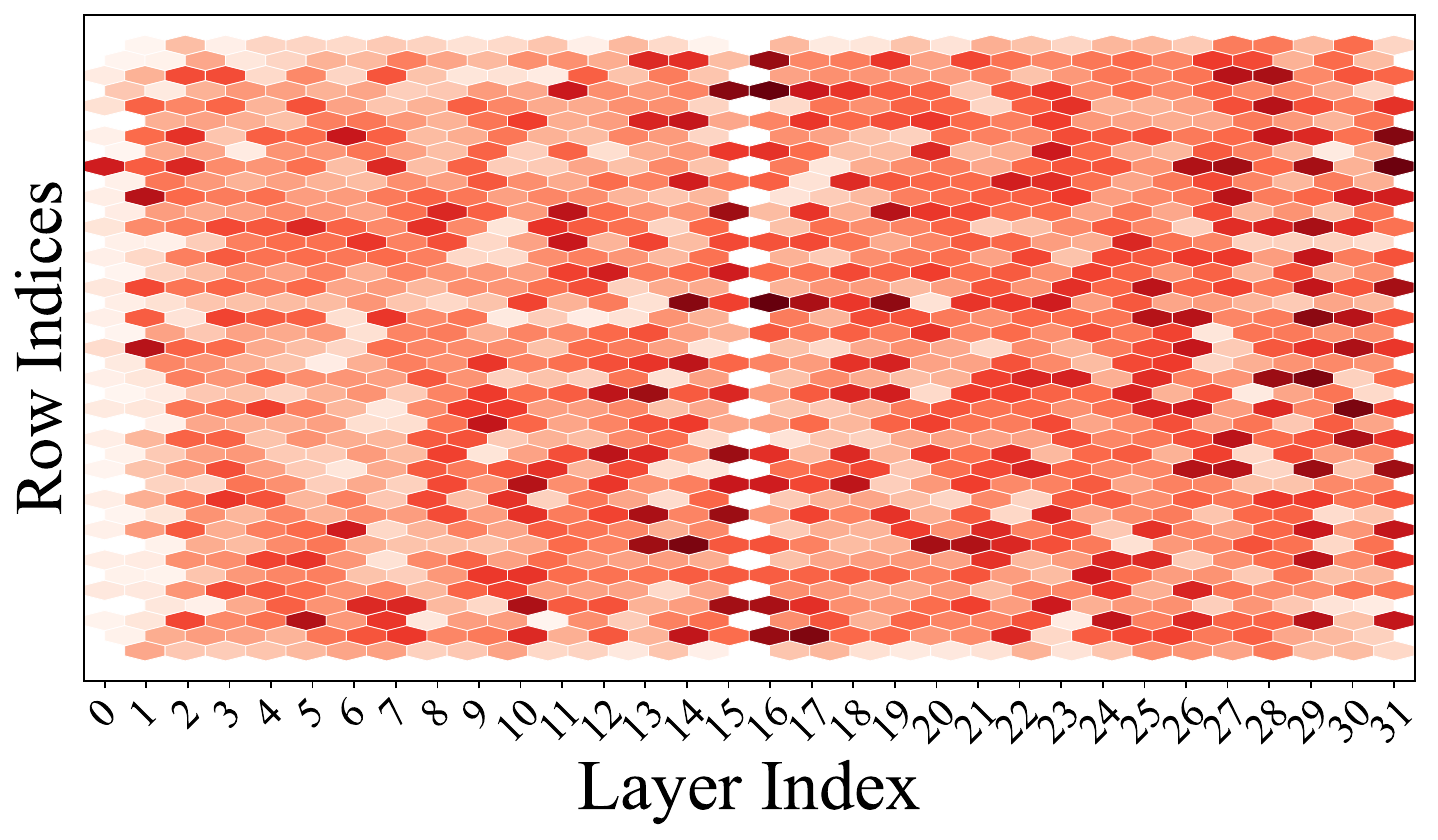}
        \caption{\texttt{v\_proj}}
        \label{fig:v_proj}
    \end{subfigure}
    \caption{Layer-wise distribution of high-activation neurons across different projection types. The heatmaps visualize indices identified for freezing (top 30\% global activation). Notably, \texttt{q\_proj} and \texttt{k\_proj} show high concentration in early layers, whereas \texttt{v\_proj} exhibits a more uniform distribution, suggesting value information is distributed more evenly across the depth of the network.}
    \label{fig:activation_heatmaps}
\end{figure*}

Since these salient neurons characterize the core upstream knowledge, we freeze their corresponding parameters in the weight matrix. Specifically, we construct a binary gradient mask $\mathbf{M}_{grad} \in \{0, 1\}^{d_{out} \times d_{in}}$ such that:
\begin{equation}
\begin{aligned}
(M_{grad})_{k,:} &= \mathbf{0} \quad \text{if } k \in \mathcal{I}, \\
(M_{grad})_{k,:} &= \mathbf{1} \quad \text{otherwise} \\
M_{ret} &= \mathbf{1} - M_{grad}
\end{aligned}
\label{eq:mask}
\end{equation}
where $\mathbf{0}$ and $\mathbf{1}$ are row vectors of size $d_{in}$. During downstream optimization, the parameters associated with salient regions are held constant via masked gradient updates:
\begin{equation}
\mathbf{W}^{(t+1)} = \mathbf{W}^{(t)} - \eta \cdot \left( \nabla_{\mathbf{W}} \mathcal{L}_{down} \odot \mathbf{M}_{grad} \right)
\label{eq:update}
\end{equation}
where $\eta$ is the learning rate and $\odot$ denotes the Hadamard product. In practice, the PyTorch implementation achieves this by utilizing a training-time wrapper, \texttt{AwareLinear}, which decomposes the linear layer into active and frozen components to efficiently manage gradient updates without altering the underlying operator's logic. This wrapper is removed post-training, leaving the original model configuration intact. This row-wise freezing mechanism ensures that the most influential output features of the pre-trained LLaVA model remain intact while allowing less critical regions to adapt to new tasks. As shown in \cref{fig:activation_heatmaps}, the distribution of these salient neurons varies across transformer layers and projection types.

\subsection{Selection of Target Linear}
\label{sec:target_linear}
We specifically apply AWARe to the linear projection layers within the self-attention mechanism (i.e., \texttt{q\_proj}, \texttt{k\_proj}, \texttt{v\_proj}) and the linear layers within the multimodal projector (\texttt{mm\_projector}). This selection is informed by recent research \cite{zhu2025teachlargemultimodalmodels} which demonstrates that updating self-attention projections tends to cause significantly less catastrophic forgetting of pre-existing knowledge compared to the multilayer perceptron (MLP) blocks. By focusing our retention strategy on those linear, we balance the preservation of core cross-modal reasoning capabilities with the flexibility needed for downstream task adaptation. An ablation analysis is provided in \cref{sec:ablation_studies}.
Consistent with this finding, all other components of the LLM, including MLP blocks and the output projection ($\mathbf{W}_o$), are kept entirely frozen during adaptation. This configuration not only maximizes parameter efficiency but also minimizes the risk of distorting the pre-trained feature space in regions less critical for downstream alignment.
The complete procedure for AWARe is summarized in \cref{alg:aware}.

\begin{algorithm}[t]
\caption{AWARe}
\label{alg:aware}
\begin{algorithmic}[1]
\INPUT Pre-trained model $\mathcal{M}_\theta$, Upstream task $\mathcal{T}_{up}$, Downstream task $\mathcal{T}_{down}$, Retention ratio $\rho$
\OUTPUT Fine-tuned model $\mathcal{M}_\theta$
\STATE \textbf{Phase 1: Knowledge Profiling}
\STATE Sample calibration set $\mathcal{S}_{cal} \sim \mathcal{T}_{up}$
\FOR{each target layer $l$ in $\mathcal{M}_\theta$}
    \STATE Perform forward pass on $\mathcal{S}_{cal}$ to obtain activations $\mathbf{H}^{(l)}$
    \STATE Calculate neuron saliency $\mathbf{s}^{(l)}$ using \cref{eq:norm,eq:normalize,eq:saliency}
\ENDFOR
\STATE $\mathcal{I} \leftarrow \text{GlobalTopIndices}(\{\mathbf{s}^{(l)}\}_l, \rho)$
\FOR{each target layer $l$ in $\mathcal{M}_\theta$}
    \STATE Extract layer-specific retained indices $\mathcal{I}^{(l)}$ from $\mathcal{I}$
    \STATE Generate row-wise gradient mask $\mathbf{M}_{grad}^{(l)}$ via \cref{eq:mask}
\ENDFOR
\STATE \textbf{Phase 2: Constrained Fine-tuning}
\WHILE{not converged on $\mathcal{T}_{down}$}
    \STATE $\mathcal{L} \leftarrow \text{ComputeLoss}(\mathcal{M}_\theta, \mathcal{T}_{down})$
    \STATE Update parameters: $\theta \leftarrow \theta - \eta \cdot (\nabla_\theta \mathcal{L} \odot \mathbf{M}_{grad})$
\ENDWHILE
\end{algorithmic}
\end{algorithm}

%% file: sections/4_Experiments.tex
\input{tables/Table1}

\input{tables/Table7}

In this section, we evaluate the effectiveness of our proposed method by comparing it with several state-of-the-art baselines on multimodal downstream tasks. We also conduct extensive ablation studies to analyze the impact of key components and hyperparameters.

\subsection{Experimental Setup}

\textbf{Datasets and Tasks.} We use LLaVA-v1.5-7b \cite{LLaVA15_CVPR24} as the base model for training. The model has been pre-trained on several upstream tasks, including OKVQA \cite{OKVQA_CVPR19}, OCRVQA \cite{OCRVQA_ICDAR19}, GQA \cite{GQA_CVPR19}, and TextVQA \cite{TextVQA_CVPR19}. For downstream evaluation, we use IconQA \cite{IconQA_NeurIPS21} for VQA and COCO-Caption \cite{COCO_ECCV14} for image captioning, where captioning performance is measured by CIDEr. For continual instruction tuning, we evaluate LLaVA-v1.5-7b on MLLM-DCL \cite{MLLM_CL_arXiv25}, a five-task sequence covering remote sensing (RS), medical VQA (Med), autonomous driving (AD), science reasoning (Sci), and financial understanding (Fin). We provide additional experiments on Qwen2.5-VL-7B \cite{Qwen2_5VL_arXiv25} in \cref{sec:qwen25vl_mllm_dcl}.

\textbf{Implementation Details.} During training, we freeze the vision tower to preserve the pre-trained visual representations. The complete training configuration and hyperparameter settings are provided in \cref{sec:exp_config}.

\subsection{Baselines}

We evaluate AWARe under two complementary settings. First, in single-task downstream adaptation, the model learns one new downstream task and is evaluated for both target-task learning and forgetting on upstream tasks. In this setting, we compare against established parameter-efficient fine-tuning and anti-forgetting baselines, including \textbf{Full-FT}, \textbf{LoRA} \cite{LoRA_ICLR22}, \textbf{DoRA} \cite{DoRA_ICML24}, \textbf{DARE} \cite{DARE_ICML24}, \textbf{Orth-Reg} \cite{Orth-Reg_ECCV24}, \textbf{Model Tailor} \cite{ModelTailor_ICML24}, \textbf{LoRASculpt} \cite{LORASCULPT_CVPR25}, and \textbf{SPIDER} \cite{SPIDER_ICML25}. Second, in continual instruction tuning, the model learns a sequence of subtasks and is evaluated after each training stage. For this setting, we compare against continual learning baselines including \textbf{ModalPrompt} \cite{ModalPrompt_EMNLP25}, \textbf{SEFE} \cite{SEFE_ICML25}, \textbf{HiDe-LLaVA} \cite{HiDeLLaVA_ACL25}, \textbf{CL-MoE} \cite{CLMoE_CVPR25}, \textbf{O-LoRA} \cite{OLoRA_EMNLP23}, \textbf{MoELoRA} \cite{CoIN_NeurIPS24}, and \textbf{DISCO} \cite{DISCO_ICCV25}.
Detailed descriptions for all baselines are provided in \cref{sec:baseline_details}.

\subsection{Single-Task Downstream Adaptation}
\label{sec:main_results}

We evaluate the performance of all methods by decomposing continual learning capabilities into two core components: \textbf{Knowledge Retention ($\mathcal{R}$)} and \textbf{Learning Efficiency ($\mathcal{E}$)}. To provide a balanced assessment, we report their Harmonic Mean ($\mathcal{H}$):
\begin{equation}
\mathcal{H} = \frac{2 \cdot \mathcal{R} \cdot \mathcal{E}}{\mathcal{R} + \mathcal{E}}
\end{equation}
The components are defined as follows:
\begin{fullitemize}
    \item \textbf{$\mathcal{R}$:} A stability metric measuring the preservation of performance on upstream tasks after fine-tuning. It is calculated as the average ratio of the fine-tuned accuracy to the base (zero-shot) accuracy:
    \begin{equation}
    \mathcal{R} = \frac{1}{|\mathcal{T}_{up}|} \sum_{i \in \mathcal{T}_{up}} \frac{Acc_{i, \text{final}}}{Acc_{i, \text{base}}}
    \end{equation}
    where $Acc_{i, \text{base}}$ is the model's initial performance before fine-tuning. A sharp drop in $\mathcal{R}$ indicates catastrophic forgetting.
    \item \textbf{$\mathcal{E}$:} A plasticity metric measuring the model's adaptation to the new downstream task. To normalize against model capacity, it is defined relative to the Full Fine-Tuning (Full-FT) baseline:
    \begin{equation}
    \mathcal{E} = \frac{Acc_{down, \text{Method}}}{Acc_{down, \text{Full-FT}}}
    \end{equation}
    where $Acc_{down, \text{Full-FT}}$ represents the performance upper bound achievable by Full-FT.
\end{fullitemize}

As shown in \cref{tab:main_results}, our method achieves a superior harmonic balance $\mathcal{H}$ between plasticity and stability. Compared to other methods, our approach maintains high retention rates $\mathcal{R}$ while achieving efficiency $\mathcal{E}$. Notably, when upstream task data is restricted, we used the MMMU general dataset as a calibration dataset and still achieved relatively good results, outperforming both \lorasculpt{} and \spider{}. We further verify that AWARe is not limited to the LLaVA-v1.5 backbone by conducting an additional Qwen2.5-VL base-model experiment; these results are reported in \cref{sec:qwen25vl_mllm_dcl}.

\subsection{Continual Instruction Tuning on MLLM-DCL}
\label{sec:mllm_dcl}

To further demonstrate the effectiveness of our proposed method in sequential learning scenarios, we evaluate AWARe on the MLLM-DCL benchmark \cite{MLLM_CL_arXiv25}. This benchmark is specifically designed to assess continual instruction tuning for multimodal large language models. As shown in \cref{tab:mllm_dcl}, AWARe achieves the highest average per-task performance (\textit{Avg}) and the best final performance after the last task (\textit{Last}), improving the overall averages by 3.18 and 1.77 points over the strongest baseline, respectively. The detailed CL results are provided in \cref{sec:mllm_dcl_details}. Overall, AWARe effectively balances plasticity and stability, confirming its superiority in continual instruction tuning settings.

\subsection{Ablation Studies}
\label{sec:ablation_studies}

We conduct ablation studies to investigate the contribution of different components and hyperparameter choices in our framework.

\textbf{Effectiveness of Activation-based Selection.} We compare our activation-based parameter selection strategy against weight norm-based selection, a hybrid strategy ($0.3 \times \text{Weight} + 0.7 \times \text{Activation}$), and a baseline using random parameter selection at the same ratio (30\%). As shown in \cref{tab:ablation_selection}, random selection significantly fails to protect the upstream knowledge, with the average performance on upstream tasks (\textit{Source Avg}) dropping to 49.94 on the IconQA task. In contrast, our activation-based selection consistently achieves the best balance, outperforming both random and weight-based selection in both \textit{Source Avg} and \textit{Target} scores. 

\input{tables/Table2}

\textbf{Impact of Selection Ratio and Strategy.} We explore the effect of varying the selection ratio (from 1\% to 90\%) and comparing Layer-Balanced versus Global-Highest selection strategies. As shown in \cref{tab:ablation_ratio}, the Global-Highest selection strategy at a 30\% ratio achieves the optimal balance between stability and plasticity, yielding the highest harmonic mean ($\mathcal{H}=103.2$) and superior knowledge retention ($\mathcal{R}=98.4$). 

\input{tables/Table3}

\textbf{Calibration Dataset Composition.} We examine how the composition of the calibration dataset influences the model's resistance to catastrophic forgetting on the IconQA task. As shown in \cref{tab:ablation_calibration}, using any individual upstream task (e.g., OKVQA, OCRVQA, GQA, or TextVQA) as the calibration source consistently preserves knowledge, with relatively small performance variations across different configurations. However, our comprehensive calibration approach (ALL), which integrates samples from all upstream sources, yields the most balanced and superior overall performance ($\mathcal{H}=103.2$). This demonstrates that while AWARe is robust to the specific choice of calibration task, a diverse mixture of upstream data ensures the most effective protection across the model's entire functional landscape.

\input{tables/Table4}

\begin{figure}[h]
    \centering
    \includegraphics[width=\columnwidth]{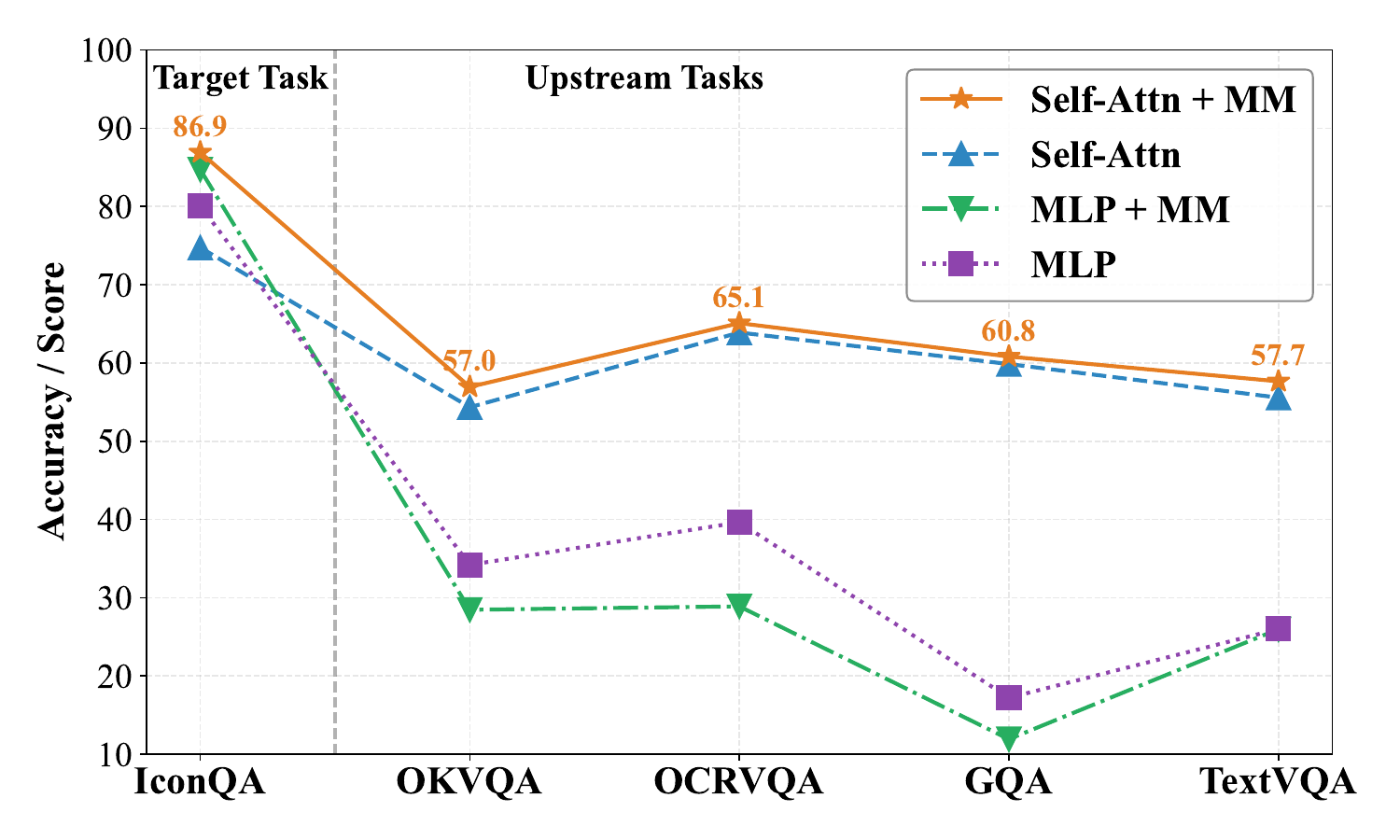}
    \caption{\textbf{Ablation Study on Training Target.} Performance comparison across different architecture components on the IconQA and upstream tasks. Applying AWARe to both \texttt{self-attn} and \texttt{mm\_projector} achieves the best stable-plastic trade-off.}
    \label{fig:training_target_ablation}
\end{figure}

\textbf{Training Targets.} We investigate the impact of applying our method to different components within the model architecture. As illustrated in \cref{fig:training_target_ablation}, targeting both the self-attention projections and the multimodal (\texttt{mm\_projector}) yields the most favorable balance between plasticity and stability. Specifically, the \texttt{mm\_projector} is essential for learning new downstream concepts; Locking it limits adaptation and causes a roughly 10-point decline in target performance. Furthermore, we observe that fine-tuning \texttt{MLP} blocks results in a severe performance collapse on upstream tasks (dropping below 30-point), even with our adaptive protection. This reinforces the hypothesis that MLP weights house the bulk of the model's fundamental general knowledge, whereas self-attention projections provide a more flexible landscape for specialized task adaptation.

\subsection{Sensitivity to Calibration Datasets}
\label{sec:calibration_dataset_sensitivity}
To assess the statistical significance and robustness of our results, we evaluate the sensitivity of AWARe to the randomness of the calibration set. We repeat the Knowledge Profiling phase over three independent runs with different random seeds for sample selection. As shown in \cref{tab:ablation_sampling}, the performance metrics exhibit minimal standard deviation across runs, demonstrating that the identified salient neurons are consistent and statistically stable artifacts of the model structure, rather than noise from specific data samples. Moreover, increasing the calibration size from 200 to 400 samples yields negligible performance differences, confirming that 200 samples are statistically sufficient to robustly estimate parameter importance.
\input{tables/Table6}

%% file: tables/Table1.tex
\begin{table*}[t]\small
\centering
\scriptsize{
\resizebox{\linewidth}{!}{
\setlength\tabcolsep{3.pt}
\renewcommand\arraystretch{1.1}
\begin{tabular}{r||cccc|ccc|cIcccc|ccc|c}
\hline\thickhline
\rowcolor{gray!20}
 & 
\multicolumn{8}{cI}{\bm{\iconqa{}}} & 
\multicolumn{8}{c}{\bm{{\coco{}}}}
\\
\cline{2-17} 
\rowcolor{gray!20}
\multirow{-2}{*}{Methods\quad}  
& \okvqa{} & \ocrvqa{} & \gqa{} & \textvqa{} & \textit{Target} & $\mathcal{R}$ & $\mathcal{E}$  & $\mathcal{H}$
& \okvqa{} & \ocrvqa{} & \gqa{} & \textvqa{} & \textit{Target} & $\mathcal{R}$ & $\mathcal{E}$  & $\mathcal{H}$
\\
\hline\hline

\zeroshot{} 
& 57.99 & 66.20 & 61.93 & 58.23 & 23.18 & 100.0 & 28.9  & 44.9 
& 57.99 & 66.20 & 61.93 & 58.23 & 40.40 & 100.0 & 42.3  & 59.4 
\\
\hline\hline

\rowcolor{gray!10} \fullft{} \dag
& 0.04 & 0.00 & 0.00 & 0.46 & 80.15 & 0.2 & 100.0 & 0.4  
& 0.00 & 0.00 & 0.00 & 0.04 & 95.59 & 0.0 & 100.0 & 0.0 
\\

\lora{} 
& 45.06 & 48.40 & 49.22 & 36.47 & 63.65 & 73.3 & 79.4 & 76.2
& 44.92 & 59.90 & 54.88 & 39.75 & 110.27 & 81.6 & 115.4 & 95.6 
\\

\rowcolor{gray!10} \dora{} 
& 49.94 & 54.40 & 55.39 & 47.26 & 84.48 & 84.7 & 105.4 & 93.9 
& 45.38 & 59.75 & 55.40 & 41.38 & 112.60 & 82.6 & 117.8 & 97.1 
\\

\orthreg{} 
& 53.12 & 56.10 & 57.43 & 51.00 & 84.52 & 89.1 & 105.5 & 96.6 
& 48.11 & 59.85 & 56.83 & 46.69 & 111.87 & 86.5 & 117.0 & 99.5 
\\

\rowcolor{gray!10} \dare{}
& 44.39 & 46.85 & 48.75 & 34.84 & 82.28 & 71.5 & 102.7 & 84.3  
& 42.98 & 58.65 & 53.92 & 38.61 & 108.57 & 79.5 & 113.6 & 93.5
\\

\tailor{} 
& 50.07 & 53.85 & 53.04 & 46.92 & 81.76 & 83.4 & 102.0 & 91.8  
& \underline{50.99} & 60.60 & \underline{58.54} & 49.65 & 117.64 & \textbf{89.9} & 123.1 & 103.9 
\\

\rowcolor{gray!10} \lorasculpt{}
& \underline{53.52} & 59.50 & \underline{57.63} & \underline{53.76} & 85.26 & \underline{91.8} & 106.4 & \underline{98.6}  
& 49.99 & 58.65 & 57.63 & \underline{50.73} & \underline{120.35} & 88.8 & \underline{125.9} & \underline{104.2}
\\

\spider{}
& 50.41 & \underline{61.49} & 56.54 & 51.54 & \underline{85.29} & 90.0 & \underline{106.4} & 97.5 
& 49.50 & \textbf{65.00} & 58.14 & 45.33 & 114.74 & 89.2 & 120.0 & 102.3
\\

\hline
\rowcolor{LightOrange}
\textbf{\ourmethod{}}
& \textbf{56.95} & \textbf{65.10} & \textbf{60.83} & \textbf{57.66} & \textbf{86.92} & \textbf{98.4} & \textbf{108.4} & \quad\textbf{103.2}\greenup{4.6}\quad   
& \textbf{53.01} & \underline{63.45} & \textbf{60.32} & \textbf{53.46} & \textbf{120.94} & \textbf{94.2} & \textbf{126.5} & \quad\textbf{108.0}\greenup{3.8}\quad  
\\

\hline
\rowcolor{LightOrange!30}
\textit{\ourmethod{} (MMMU)}
& 56.86 & 63.35 & 60.93 & 56.77 & 85.23 & 97.4 & 106.3 & \quad101.7\greenup{3.1}\quad   
& 53.25 & 63.12 & 60.38 & 52.74 & 121.22 & 93.9 & 126.8 & \quad107.9\greenup{3.7}\quad  
\\
\hline

\end{tabular}}}
\captionsetup{font=small}
\caption{{
\textbf{Comparison with State-of-the-Art Fine-Tuning Solutions on LLaVA-v1.5-7B} for \iconqa{} and \coco{}.
The optimal and sub-optimal results are denoted by boldface and underlining. {\greenup{}} means improved accuracy compared with the sub-optimal results.
\dag Specifically, the near-zero performance of \fullft{} on upstream tasks reflects extreme catastrophic forgetting due to overfitting to the target task.
More details about the metrics are provided in \cref{sec:main_results}. Training setting are detailed in \cref{sec:exp_config}.
}}
\label{tab:main_results}
\vspace{-10pt}
\end{table*}

%% file: tables/Table7.tex
\begin{table*}[t]\small
\centering
\scriptsize{
\resizebox{\linewidth}{!}{
\setlength\tabcolsep{3.pt}
\renewcommand\arraystretch{1.1}
\begin{tabular}{l||ccccc|c||ccccc|c}
\hline\thickhline
\rowcolor{gray!20}
 & \multicolumn{6}{c||}{Avg} & \multicolumn{6}{c}{Last} \\
\rowcolor{gray!20}
\multirow{-2}{*}{Method} & RS $\uparrow$ & Med $\uparrow$ & AD $\uparrow$ & Sci $\uparrow$ & Fin $\uparrow$ & Average $\uparrow$ & RS $\uparrow$ & Med $\uparrow$ & AD $\uparrow$ & Sci $\uparrow$ & Fin $\uparrow$ & Average $\uparrow$ \\
\hline\hline
ModalPrompt & 53.19 & 45.73 & 40.78 & 41.82 & 87.82 & 53.87 & 53.63 & 45.68 & 40.77 & 41.81 & 87.82 & 53.94 \\
\rowcolor{gray!10}
HiDe & 75.41 & 48.71 & 37.06 & 42.49 & 81.55 & 57.04 & 74.31 & \underline{48.95} & 33.21 & 38.54 & 81.55 & 55.31 \\
LoRA-FT & 73.34 & 47.97 & 38.68 & 45.16 & 87.45 & 58.52 & 69.65 & 41.59 & 25.43 & 40.88 & 87.45 & 53.00 \\
\rowcolor{gray!10}
MoELoRA & 77.19 & 47.02 & 37.30 & 44.40 & 86.75 & 58.53 & \underline{77.54} & 41.85 & 27.62 & 40.13 & 86.75 & 54.78 \\
CL-MoE & 73.51 & 49.97 & 38.55 & 45.73 & 88.74 & 59.30 & 71.34 & 46.84 & 26.33 & 41.17 & 88.74 & 54.88 \\
\rowcolor{gray!10}
O-LoRA & 76.12 & 47.38 & 40.97 & 46.02 & 87.15 & 59.53 & 74.64 & 44.42 & 30.02 & 41.47 & 87.15 & 55.54 \\
SEFE & 77.38 & \underline{51.41} & 44.24 & 44.95 & 86.82 & 60.96 & 77.26 & \textbf{50.37} & 37.21 & 40.87 & 86.82 & 58.51 \\
\rowcolor{gray!10}
DISCO & \underline{77.41} & 48.04 & \underline{48.70} & \underline{48.65} & \underline{89.22} & \underline{62.40} & 76.49 & 44.48 & \underline{44.84} & \textbf{46.61} & \underline{89.22} & \underline{60.33} \\
\rowcolor{LightOrange}
\textbf{AWARe} & \textbf{79.88} & \textbf{55.42} & \textbf{51.38} & \textbf{49.79} & \textbf{91.45} & \textbf{65.58}\greenup{3.18} & \textbf{78.96} & 47.01 & \textbf{46.94} & \underline{46.13} & \textbf{91.45} & \textbf{62.10}\greenup{1.77} \\
\hline\thickhline
\end{tabular}}}
\captionsetup{font=small}
\caption{{
\textbf{Detailed continuous instruction tuning performance on the MLLM-DCL benchmark} \cite{MLLM_CL_arXiv25}. \textit{Avg} reports the average score for each subtask across all later evaluation points after that subtask is learned, while \textit{Last} reports the final score on each subtask after training on the last subtask.
}}
\label{tab:mllm_dcl}
\vspace{-10pt}
\end{table*}

%% file: tables/Table2.tex
\begin{table}[h]\small
\centering
\scriptsize{
\resizebox{\linewidth}{!}{
\setlength\tabcolsep{3.pt}
\renewcommand\arraystretch{1.1}
\begin{tabular}{l||ccIcc}
\hline\thickhline
\rowcolor{gray!20}
 & 
\multicolumn{2}{cI}{\bm{\iconqa{}}} & 
\multicolumn{2}{c}{\bm{{\coco{}}}}
\\
\cline{2-5}
\rowcolor{gray!20}
\multirow{-2}{*}{Strategy\quad}  
& \textit{Source Avg} & \textit{Target}
& \textit{Source Avg} & \textit{Target}
\\
\hline\hline

Random Selection
& 49.94 & \underline{86.32}
& 49.46 & \underline{116.31}
\\

\rowcolor{gray!10} Weight Norm
& 58.51 & 85.51
& 53.52 & 115.19
\\

$0.3W + 0.7A$
& \underline{58.69} & 84.66
& \underline{53.90} & 115.24 
\\

\rowcolor{LightOrange}
\textbf{Activation}
& \textbf{60.13} & \textbf{86.92}
& \textbf{57.56} & \textbf{120.94}
\\
\hline

\end{tabular}}}
\captionsetup{font=small}
\caption{{
\textbf{Ablation Study on Parameter Selection Strategy.} We compare different methods for selecting parameters to retain: random selection (Random Selection), weight magnitude (Weight Norm), activation magnitude (Activation), and a hybrid approach ($0.3W + 0.7A$).
}}
\label{tab:ablation_selection}
\vspace{-10pt}
\end{table}

%% file: tables/Table3.tex
\begin{table}[h]\small
\centering
\scriptsize{
\resizebox{\linewidth}{!}{
\setlength\tabcolsep{3.pt}
\renewcommand\arraystretch{1.1}
\begin{tabular}{c||cccc|c|ccc}
\hline\thickhline
\rowcolor{gray!20}
Ratio & \okvqa{} & \ocrvqa{} & \gqa{} & \textvqa{} & \textit{Target} & $\mathcal{R}$ & $\mathcal{E}$  & $\mathcal{H}$ \\
\hline\hline

\multicolumn{9}{l}{\textcolor{gray}{\textit{Layer balanced selection}}} \\
\hline

1\% 
& 55.33 & 64.00 & 60.17 & 54.39 & 82.44 & 95.7 & 102.9 & 99.2 \\

\rowcolor{gray!10}
10\%
& 55.33 & \underline{65.20} & 59.64 & 55.16 & \underline{86.84} & 96.3 & \underline{108.3} & 102.0 \\

30\% 
& 56.28 & 62.75 & 60.62 & 56.13 & 85.29 & 96.5 & 106.4 & 101.2 \\

\rowcolor{gray!10}
40\% 
& 56.47 & 63.85 & 60.46 & 56.94 & 86.13 & 97.3 & 107.5 & 102.1 \\

50\% 
& 57.33 & 63.95 & 60.80 & 57.19 & 86.21 & 97.9 & 107.6 & 102.5 \\

\rowcolor{gray!10}
70\% 
& 57.69 & 64.20 & 61.77 & 57.54 & 85.67 & 98.7 & 106.9 & 102.6 \\

90\% 
& 57.65 & 63.70 & \underline{62.01} & \underline{57.92} & 84.07 & 98.7 & 104.9 & 101.7 \\

\hline
\multicolumn{9}{l}{\textcolor{gray}{\textit{Global highest selection}}} \\
\hline

1\% 
& 55.26 & 63.40 & 59.89 & 54.96 & 82.65 & 95.6 & 103.1 & 99.2 \\

\rowcolor{gray!10}
10\%
& 55.88 & 63.25 & 59.59 & 55.39 & 86.13 & 95.8 & 107.5 & 101.3 \\

\rowcolor{LightOrange}
\textbf{30\%} 
& 56.95 & 65.10 & 60.83 & 57.66 & \textbf{86.92} & 98.4 & \textbf{108.4} & \textbf{103.2} \\

\rowcolor{gray!10}
40\% 
& 57.66 & 64.35 & 61.47 & 57.77 & 86.64 & 98.7 & 108.1 & \textbf{103.2} \\

50\% 
& 57.71 & 64.75 & 61.87 & 57.86 & 85.97 & \underline{99.1} & 107.3 & \underline{103.0} \\

\rowcolor{gray!10}
70\% 
& \underline{57.78} & 64.35 & \textbf{62.03} & 57.84 & 80.64 & 99.0 & 100.6 & 99.8 \\

90\% 
& \textbf{58.13} & \textbf{65.55} & 61.89 & \textbf{58.33} & 54.62 & \textbf{99.8} & 68.1 & 81.0 \\
				
\hline
\end{tabular}}}
\captionsetup{font=small}
\caption{{
\textbf{Ablation Study on Selection Ratio.} We evaluate the performance of our method with different ratios of parameters selected for retention on the \iconqa{} task.
}}
\label{tab:ablation_ratio}
\vspace{-10pt}
\end{table}

%% file: tables/Table4.tex
\begin{table}[h]\small
\centering
\scriptsize{
\resizebox{\linewidth}{!}{
\setlength\tabcolsep{3.pt}
\renewcommand\arraystretch{1.1}
\begin{tabular}{l||cccc|c|ccc}
\hline\thickhline
\rowcolor{gray!20}
Calibration Task & \okvqa{} & \ocrvqa{} & \gqa{} & \textvqa{} & \textit{Target} & $\mathcal{R}$ & $\mathcal{E}$  & $\mathcal{H}$ \\
\hline\hline

\okvqa{} 
& 55.17 & 63.96 & 58.56 & 54.72 & \textbf{86.97} & 95.1 & \textbf{108.5} & 101.4 \\

\rowcolor{gray!10} \ocrvqa{} 
& 54.76 & \textbf{65.16} & 58.28 & 54.45 & 85.85 & 95.2 & 107.1 & 100.8 \\

\gqa{}
& \underline{55.92} & 64.22 & \underline{60.12} & 54.97 & 86.21 & 96.3 & 107.6 & \underline{101.6} \\

\rowcolor{gray!10} \textvqa{}
& 55.56 & 64.40 & 59.78 & \underline{56.11} & 85.78 & \underline{96.5} & 107.0 & 101.5 \\

\rowcolor{LightOrange}
\textbf{ALL}
& \textbf{56.95} & \underline{65.10} & \textbf{60.83} & \textbf{57.66} & \underline{86.92} & \textbf{98.4} & \underline{108.4} & \textbf{103.2} \\
\hline
\end{tabular}}}
\captionsetup{font=small}
\caption{{
\textbf{Ablation Study on Calibration Dataset Composition.} We investigate how using different subsets of upstream data for activation statistics calculation affects performance on the \iconqa{} task.
}}
\label{tab:ablation_calibration}
\vspace{-10pt}
\end{table}

%% file: tables/Table6.tex
\begin{table}[h]\small
\centering
\scriptsize{
\resizebox{\linewidth}{!}{
\setlength\tabcolsep{3.pt}
\renewcommand\arraystretch{1.1}
\begin{tabular}{l||cccc|ccc|c}
\hline\thickhline
\rowcolor{gray!20}
Calibration Size & \okvqa{} & \ocrvqa{} & \gqa{} & \textvqa{} & \textit{Target} & $\mathcal{R}$ & $\mathcal{E}$ & $\mathcal{H}$ \\
\hline\hline

200 
& 56.95 & 65.10 & 60.83 & 57.66 & 86.92 & 98.4 & 108.4 & 103.2 \\

\rowcolor{gray!10} 200 
& 56.88 & 64.45 & 61.20 & 58.04 & 86.57 & 98.5 & 108.0 & 103.0 \\

200 
& 56.49 & 64.50 & 60.49 & 58.15 & 86.58 & 98.1 & 108.0 & 102.8 \\
\hline
400 
& 56.37 & 64.38 & 60.58 & 57.77 & 86.20 & 97.9 & 107.5 & 102.5 \\

\rowcolor{gray!10} 400 
& 56.74 & 64.40 & 60.38 & 57.17 & 86.69 & 97.7 & 108.2 & 102.7 \\

400 
& 57.04 & 64.62 & 60.47 & 56.77 & 86.35 & 97.8 & 107.7 & 102.5 \\
\hline
\rowcolor{gray!10}
\textit{Std. Dev.} & \textit{0.24} & \textit{0.25} & \textit{0.28} & \textit{0.48} & \textit{0.23} & \textit{0.30} & \textit{0.30} & \textit{0.30} \\
\hline
\end{tabular}}}
\captionsetup{font=small}
\caption{{
\textbf{Sensitivity Analysis on Calibration Datasets.} We evaluate the performance of AWARe on IconQA task across different calibration set sizes (200 and 400) and random sample 3 times.
}}
\label{tab:ablation_sampling}
\vspace{-10pt}
\end{table}

%% file: sections/5_Conclusion.tex
In this paper, we introduce AWARe, an activation-weighted framework for mitigating catastrophic forgetting in Multimodal Large Language Models. AWARe first profiles task-induced activations to identify salient neurons, then freezes these regions during downstream fine-tuning to preserve pre-trained capabilities. This design is architecture-agnostic, requires no additional adapter modules, and updates only a small fraction of parameters. Across downstream adaptation and MLLM-DCL experiments, AWARe consistently improves the stability-plasticity trade-off, and remains effective even when using a general-purpose calibration set such as MMMU.

%% file: sections/Appendix.tex
\section{MLLM-DCL Evaluation Details}
\label{sec:mllm_dcl_details}

We follow the MLLM-DCL benchmark \cite{MLLM_CL_arXiv25} and report four standard metrics for continuous instruction tuning. Let $T$ be the total number of sequential tasks, and let $A_{t,i}$ denote the test accuracy on task $i$ after the model has finished training on task $t$.

\begin{itemize}
    \item \textbf{Mean Fine-tune Accuracy (MFT).} MFT measures plasticity, i.e., how well the model adapts to each task immediately after learning it:
    \begin{equation}
        \mathrm{MFT} = \frac{1}{T}\sum_{i=1}^{T} A_{i,i}.
    \end{equation}
    A higher MFT indicates stronger immediate adaptation to the downstream distribution of each new task.

    \item \textbf{Mean Final Accuracy (MFN).} MFN measures the model's average performance on all tasks after the entire training sequence:
    \begin{equation}
        \mathrm{MFN} = \frac{1}{T}\sum_{i=1}^{T} A_{T,i}.
    \end{equation}
    Unlike MFT, MFN accounts for later performance degradation, so a high MFN requires both learning new tasks and retaining previous ones.

    \item \textbf{Mean Average Accuracy (MAA).} MAA summarizes the accuracy trajectory throughout continual learning by averaging the cumulative accuracy over observed tasks at each training step:
    \begin{equation}
        \mathrm{MAA} = \frac{1}{T}\sum_{t=1}^{T}\left(\frac{1}{t}\sum_{i=1}^{t} A_{t,i}\right).
    \end{equation}
    MAA rewards methods that maintain consistently high accuracy across all known tasks during the learning process.

    \item \textbf{Backward Transfer (BWT).} BWT quantifies the average final change in performance on previously learned tasks relative to their accuracy immediately after learning:
    \begin{equation}
        \mathrm{BWT} = \frac{1}{T-1}\sum_{i=1}^{T-1}\left(A_{T,i} - A_{i,i}\right).
    \end{equation}
    A less negative or positive BWT indicates less forgetting, while a negative BWT reflects degradation on earlier tasks.
\end{itemize}

\begin{table}[t]\small
\centering
\scriptsize{
\resizebox{\columnwidth}{!}{
\setlength\tabcolsep{3.pt}
\renewcommand\arraystretch{1.1}
\begin{tabular}{l||cccc}
\hline\thickhline
\rowcolor{gray!20}
Method & MFT $\uparrow$ & MFN $\uparrow$ & MAA $\uparrow$ & BWT $\uparrow$ \\
\hline\hline
ModalPrompt & 53.87 & 53.94 & 49.67 & \textbf{+0.09} \\
\rowcolor{gray!10}
HiDe & 60.77 & 55.31 & 60.68 & -6.82 \\
LoRA-FT & 64.98 & 53.00 & 61.13 & -14.97 \\
\rowcolor{gray!10}
MoELoRA & 64.94 & 54.78 & 61.76 & -12.70 \\
CL-MoE & \underline{66.06} & 54.88 & 61.79 & -13.96 \\
\rowcolor{gray!10}
O-LoRA & 65.16 & 55.54 & 62.12 & -12.03 \\
SEFE & 65.01 & 58.51 & 63.63 & -8.13 \\
\rowcolor{gray!10}
DISCO & 64.78 & \underline{60.33} & \underline{63.93} & \underline{-5.57} \\
\rowcolor{LightOrange}
\textbf{AWARe} & \textbf{67.49} & \textbf{62.10} & \textbf{67.28} & -6.74 \\
\hline\thickhline
\end{tabular}}}
\caption{Detailed CL metric summary on MLLM-DCL.}
\label{tab:mllm_dcl_cl_metrics}
\vspace{-10pt}
\end{table}

For the detailed AWARe run in \cref{tab:mllm_dcl_task_matrix}, we use 200 MMMU samples as the calibration dataset when training the first RS task. For each subsequent task, the calibration dataset consists of MMMU together with 100 samples from each previous task, which are used to recompute the activation distributions. Rows indicate the latest task after which the model is fine-tuned, and columns indicate evaluation tasks.

\begin{table}[t]
\centering
\small
\resizebox{\columnwidth}{!}{
\begin{tabular}{lccccc}
\thickhline
\rowcolor{gray!20}
Training Stage & RS & Med & AD & Sci & Fin \\
\hline
RS  & 78.87 & -- & -- & -- & -- \\
Med & 80.76 & 59.44 & -- & -- & -- \\
AD  & 80.80 & 59.03 & 54.25 & -- & -- \\
Sci & 80.02 & 56.20 & 52.94 & 53.45 & -- \\
Fin & 78.96 & 47.01 & 46.94 & 46.13 & 91.45 \\
\thickhline
\end{tabular}
}
\caption{Detailed AWARe accuracy results on MLLM-DCL.}
\label{tab:mllm_dcl_task_matrix}
\end{table}

\section{Additional Model Qwen2.5-VL}
\label{sec:qwen25vl_mllm_dcl}

To examine whether AWARe remains effective on a stronger recent backbone, we additionally evaluate Qwen2.5-VL \cite{Qwen2_5VL_arXiv25} on two downstream tasks from MLLM-DCL: Med and AD. We use HallusionBench \cite{Hallusionbench_CVPR24}, Micro-VQA \cite{Microvqa_CVPR25}, DocVQA \cite{Docvqa_CVPR21}, VStarBench \cite{Vstar_CVPR24}, and AI2D \cite{AI2D_ECCV16} as upstream evaluation tasks to measure retained general multimodal capabilities after task-specific tuning. The zero-shot row reports the original Qwen2.5-VL performance before downstream adaptation.

\begin{table}[t]\small
\centering
\scriptsize{
\resizebox{\linewidth}{!}{
\setlength\tabcolsep{3.pt}
\renewcommand\arraystretch{1.1}
\begin{tabular}{l||c|ccccc}
\hline\thickhline
\rowcolor{gray!20}
Method & DCL-Med $\uparrow$ & HallusionBench $\uparrow$ & Micro-VQA $\uparrow$ & DocVQA $\uparrow$ & VStarBench $\uparrow$ & AI2D $\uparrow$ \\
\hline\hline

\zeroshot{} & 0.357 & 0.770 & 0.480 & 0.947 & 0.620 & 0.740 \\
\hline\hline

\rowcolor{gray!10} SFT & \textbf{0.521} & 0.330 & 0.070 & 0.599 & 0.010 & 0.040 \\

\rowcolor{LightOrange}
\textbf{AWARe} & 0.477 & \textbf{0.760} & \textbf{0.480} & \textbf{0.913} & \textbf{0.650} & \textbf{0.730} \\

SPIDER & 0.452 & 0.670 & 0.440 & 0.875 & 0.580 & 0.710 \\

\rowcolor{gray!10} LoRASculpt & 0.465 & 0.720 & 0.460 & 0.863 & 0.560 & 0.690 \\

\hline
\end{tabular}}}
\captionsetup{font=small}
\caption{{
\textbf{Qwen2.5-VL results on the MLLM-DCL Med task.}
The downstream column reports DCL-Med adaptation performance, while the remaining columns report upstream task retention. Boldface marks the best downstream result and the best upstream retention among fine-tuned methods.
}}
\label{tab:qwen25vl_mllm_dcl_med}
\vspace{-10pt}
\end{table}

\begin{table}[t]\small
\centering
\scriptsize{
\resizebox{\linewidth}{!}{
\setlength\tabcolsep{3.pt}
\renewcommand\arraystretch{1.1}
\begin{tabular}{l||c|ccccc}
\hline\thickhline
\rowcolor{gray!20}
Method & DCL-AD $\uparrow$ & HallusionBench $\uparrow$ & Micro-VQA $\uparrow$ & DocVQA $\uparrow$ & VStarBench $\uparrow$ & AI2D $\uparrow$ \\
\hline\hline

\zeroshot{} & 0.213 & 0.770 & 0.480 & 0.947 & 0.620 & 0.740 \\
\hline\hline

\rowcolor{gray!10} SFT & \textbf{0.5595} & 0.500 & 0.390 & 0.190 & 0.240 & 0.370 \\

\rowcolor{LightOrange}
\textbf{AWARe} & 0.4629 & \textbf{0.760} & \textbf{0.440} & \textbf{0.901} & \textbf{0.600} & \textbf{0.710} \\

SPIDER & 0.432 & 0.650 & 0.390 & 0.856 & 0.540 & 0.650 \\

\rowcolor{gray!10} LoRASculpt & 0.445 & 0.690 & 0.420 & 0.847 & 0.520 & 0.690 \\

\hline
\end{tabular}}}
\captionsetup{font=small}
\caption{{
\textbf{Qwen2.5-VL results on the MLLM-DCL AD task.}
The downstream column reports DCL-AD adaptation performance, while the remaining columns report upstream task retention. Boldface marks the best downstream result and the best upstream retention among fine-tuned methods.
}}
\label{tab:qwen25vl_mllm_dcl_ad}
\vspace{-10pt}
\end{table}

Across both downstream tasks, SFT reaches the highest target-task accuracy but causes severe degradation on upstream capabilities, especially on DocVQA, VStarBench, and AI2D. In contrast, AWARe consistently preserves the strongest upstream performance among fine-tuned methods while retaining competitive downstream adaptation, indicating that activation-weighted parameter freezing transfers to Qwen2.5-VL and provides a robust stability-plasticity trade-off.

\section{Parameter Efficiency Analysis}
\label{sec:param_analysis}

In this section, we provide a detailed calculation of the trainable parameter ratio in our optimal setting. 
Consider a standard Transformer block in LLaMA architecture with hidden dimension $h$. The parameters can be categorized into Self-Attention and Feed-Forward Network (FFN) modules:

\begin{itemize}
    \item \textbf{Self-Attention:} Consists of four projection matrices: $\mathbf{W}_q, \mathbf{W}_k, \mathbf{W}_v, \mathbf{W}_o \in \mathbb{R}^{h \times h}$. The total parameter count is $4h^2$.
    \item \textbf{FFN (SwiGLU):} Comprises three matrices: $\mathbf{W}_{up}, \mathbf{W}_{gate} \in \mathbb{R}^{h_{inter} \times h}$ and $\mathbf{W}_{down} \in \mathbb{R}^{h \times h_{inter}}$. In LLaMA, the intermediate dimension is typically set to $h_{inter} \approx \frac{8}{3}h$. The total parameter count is approximately $3 \times \frac{8}{3}h \times h = 8h^2$.
\end{itemize}
Thus, the total number of parameters per block is approximately $12h^2$.

In our best experimental setting, we apply the AWARe strategy specifically to the query, key, and value projection layers ($\mathbf{W}_q, \mathbf{W}_k, \mathbf{W}_v$) with a retention ratio of $\rho = 30\%$, while freezing the remaining components. The number of trainable parameters is calculated as:
\begin{equation}
    N_{train} = 3 \times (1 - \rho) \times h^2 = 3 \times 0.7 \times h^2 = 2.1h^2
\end{equation}
Consequently, the ratio of trainable parameters to the total parameter count is:
\begin{equation}
    \text{Ratio} = \frac{2.1h^2}{12h^2} \approx 17.5\%
\end{equation}
This analysis demonstrates that our method effectively updates only a quarter of the model parameters, significantly reducing computational overhead while preserving generalization capabilities.

\section{Experimental Configuration}
\label{sec:exp_config}

We summarize the hyperparameter configuration that achieves the optimal performance reported in our main results in \cref{tab:best_hyperparams}, specifically under the \textbf{Global-Highest 30\%} setting. All experimental results are averaged over three random runs.

\begin{table*}[t]
\centering
\small
\begin{tabular}{l|l}
\toprule
\textbf{Hyperparameter} & \textbf{Value} \\
\midrule
\multicolumn{2}{l}{\textit{Knowledge Profiling Phase}} \\
\midrule
Calibration Dataset Composition & OKVQA, OCRVQA, GQA, TextVQA (200 samples each) \\
\textit{When Upstream Dataset is restricted} & MMMU (20 samples each category 600 samples in total) \\
\hline
Selection Strategy & Global Highest Activation \\
Training Target Components & \texttt{q\_proj}, \texttt{k\_proj}, \texttt{v\_proj}, \texttt{mm\_projector} \\
Retention Quota ($\rho$) & 30\% (Top 30\% salient neurons frozen) \\
\midrule
\multicolumn{2}{l}{\textit{Downstream Fine-tuning Phase}} \\
\midrule
Epochs & 3 \\
Global Batch Size & 16 \\
Learning Rate (LLM Components) & $2 \times 10^{-5}$ \\
Learning Rate (MM Projector) & $2 \times 10^{-4}$ \\
LR Scheduler & Cosine Annealing (Warmup ratio: 0.03) \\
Optimizer & AdamW \\
\bottomrule
\end{tabular}
\caption{Detailed Hyperparameter Settings for the Optimal AWARe Configuration.}
\label{tab:best_hyperparams}
\end{table*}

\section{Baseline Details}
\label{sec:baseline_details}

\textbf{Single-task downstream adaptation.} We calculate baselines using the following configurations:
\begin{fullitemize}
    \item \textbf{Full-FT}: Full fine-tuning of the model parameters.
    \item \textbf{LoRA}\pub{ICLR'22}\cite{LoRA_ICLR22}: Low-Rank Adaptation, which injects trainable rank decomposition matrices.
    \item \textbf{DoRA}\pub{ICML'24}\cite{DoRA_ICML24}: Enhances LoRA’s learning capacity and training stability by decomposing weights into magnitude and direction components.
    \item \textbf{DARE}\pub{ICML'24}\cite{DARE_ICML24}: Parameters from the fine-tuned model are randomly selected and rescaled to preserve both generalization and specialization capabilities.
    \item \textbf{Orth-Reg}\pub{ECCV'24}\cite{Orth-Reg_ECCV24}: Encourages fine-tuned features to remain orthogonal to pretrained features, thereby preserving model generalization.
    \item \textbf{Model Tailor}\pub{ICML'24}\cite{ModelTailor_ICML24}: Pre-trained parameters are preserved while a small fraction (e.g., 10\%) of fine-tuned parameters is replaced based on salience and sensitivity analysis.
    \item \textbf{LoRASculpt}\pub{CVPR'25}\cite{LORASCULPT_CVPR25}: The method prunes redundant LoRA parameters via sparse updates guided by weight importance and conflict-aware regularization.
    \item \textbf{SPIDER}\pub{ICML'25}\cite{SPIDER_ICML25}: Updates parameters only where specialization knowledge is more critical than generalization knowledge.
\end{fullitemize}

% Baseline results for certain benchmarks are directly adopted from the original evaluations reported in \lorasculpt{} \cite{LORASCULPT_CVPR25} and \spider{} \cite{SPIDER_ICML25} to ensure consistency and accuracy in our comparative analysis.

\textbf{Continual instruction tuning.} For the MLLM-DCL experiments, we compare against continual learning baselines that are designed for sequential task adaptation:
\begin{fullitemize}
    \item \textbf{LoRA-FT}\pub{ICLR'22}\cite{LoRA_ICLR22}: Sequential LoRA fine-tuning on each new subtask without an explicit forgetting-mitigation mechanism.
    \item \textbf{O-LoRA}\pub{EMNLP'23}\cite{OLoRA_EMNLP23}: Learns new tasks in orthogonal subspaces to reduce interference with previously acquired knowledge.
    \item \textbf{MoELoRA}\pub{NIPS'24}\cite{CoIN_NeurIPS24}: Uses a mixture-of-LoRA-experts design for continual instruction tuning.
    \item \textbf{ModalPrompt}\pub{EMNLP'25}\cite{ModalPrompt_EMNLP25}: Introduces dual-modality guided prompts to preserve multimodal knowledge during continual learning.
    \item \textbf{CL-MoE}\pub{CVPR'25}\cite{CLMoE_CVPR25}: Uses a dual-momentum mixture-of-experts mechanism for continual multimodal visual question answering.
    \item \textbf{HiDe-LLaVA}\pub{ACL'25}\cite{HiDeLLaVA_ACL25}: Decouples multimodal continual instruction tuning hierarchically to mitigate forgetting.
    \item \textbf{SEFE}\pub{ICML'25}\cite{SEFE_ICML25}: Separates superficial and essential forgetting to preserve important multimodal capabilities.
    \item \textbf{DISCO}\pub{ICCV'25}\cite{DISCO_ICCV25}: Applies federated continual instruction tuning for sequential instruction adaptation.
\end{fullitemize}